\documentclass[10 pt, conference]{ieeeconf}
\IEEEoverridecommandlockouts 
\usepackage[noadjust]{cite}

\usepackage{amsmath}
\usepackage{yhmath}
\usepackage{mathtools} 
\usepackage{amssymb} 
\usepackage{sidecap}
\usepackage{graphicx}
\usepackage{subcaption}
\usepackage{multirow}
\usepackage{epsfig}
\usepackage{psfrag}
\usepackage{caption}
\usepackage[super]{nth} 
\usepackage[normalem]{ulem} 
\usepackage{color}
\usepackage{xcolor} 
\usepackage{booktabs} 
\usepackage{siunitx} 
\usepackage{booktabs} 
\usepackage{colortbl} 
\usepackage{textcomp} 
\usepackage{tikz} 
\usepackage[noadjust]{cite} 
\usepackage{wasysym} 
\newcommand{\RNum}[1]{\uppercase\expandafter{\romannumeral #1\relax}} 

\newcommand{\realfield}[1]{\hbox{I \kern -.5em R}^{#1}}
\newcommand {\mb}[1]{\mathbf{#1}}
\newcommand {\bs}[1]{\boldsymbol{#1}}
\newcommand{\uvec}[1]{\hat{\mathbf{#1}}}
\newcommand{\T}{^{\mathrm{T}}}  

\newcommand{\cric}{cricothyrotomy~}

\definecolor{LightGray}{gray}{0.9}
\newcolumntype{a}{>{\columncolor{LightGray}}l}
\usepackage{color} 

\usepackage[textsize=scriptsize, bordercolor=white,backgroundcolor=gray!30,linecolor=black,colorinlistoftodos]{todonotes} 
\usepackage[normalem]{ulem} 
\usepackage{soul} 

\usepackage{amsfonts} 
\title{\LARGE \bf
Feasibility of Remote Landmark\\ Identification for Cricothyrotomy Using Robotic Palpation}
\author{Neel Shihora$^{1}$,~Rashid Yasin$^{1}$,~Ryan Walsh$^{2}$,~Nabil~Simaan$^{1}$$^{\dagger}$ 
\thanks{$\dagger$ Corresponding author}
\thanks{$^{1}$Department of Mechanical Engineering, Vanderbilt University, Nashville, TN 37235, USA
        {\tt (neel.shihora,~rashid.m.yasin,~nabil.simaan) @vanderbilt.edu}}
\thanks{$^{2}$Department of Emergency Medicine, Vanderbilt University Medical Center, Nashville, TN 37235, USA
        {\tt ryan.walsh@vumc.org}}
\thanks{This work was partly supported by NSF award \#1734461 and by Vanderbilt University funds}
}
\usepackage{multirow} 
\usepackage{multicol}
\usepackage{tabularx} 
\usepackage{footnote}
\makeatletter
\newcommand{\thickhline}{
    \noalign {\ifnum 0=`}\fi \hrule height 0.3mm
    \futurelet \reserved@a \@xhline
}
\newcolumntype{"}{@{\hskip\tabcolsep\vrule width 2pt\hskip\tabcolsep}}
\makeatother
\usepackage{textcomp} 
\makeatletter
\let\NAT@parse\undefined
\makeatother
\usepackage{hyperref} 
\usepackage{cleveref}
\usepackage{fancyhdr} 
\begin{document}
\maketitle
\thispagestyle{empty}
\thispagestyle{fancy}
\fancyhf{}
\renewcommand{\headrulewidth}{0pt}
\lhead{2021 IEEE/RSJ International Conference on Intelligent Robots and Systems (IROS). pp. 1785-1791, Accepted Version.}
\rfoot{\centering \scriptsize \copyright 2021 IEEE. Personal use of this material is permitted. Permission from IEEE must be obtained for all other uses, in any current or future media, including reprinting/republishing this material for advertising or promotional purposes, creating new collective works, for resale or redistribution to servers or lists, or reuse of any copyrighted component of this work in other works.}
\pagestyle{empty}
\begin{abstract}
Cricothyrotomy is a life-saving emergency intervention that secures an alternate airway route after a neck injury or obstruction. The procedure starts with identifying the correct location (the cricothyroid membrane) for creating an incision to insert an endotracheal tube. This location is determined using a combination of visual and palpation cues. Enabling robot-assisted remote cricothyrotomy may extend this life-saving procedure to injured soldiers or patients who may not be readily accessible for on-site intervention during  search-and-rescue scenarios. As a first step towards achieving this goal, this paper explores the feasibility of palpation-assisted landmark identification for cricothyrotomy. Using a cricothyrotomy training simulator, we explore several alternatives for in-situ remote localization of the cricothyroid membrane. These alternatives include a) unaided telemanipulation, b)  telemanipulation with direct force feedback, c) telemanipulation with superimposed motion excitation for on-line stiffness estimation and display, and d) fully autonomous palpation scan initialized based on the user's understanding of key anatomical landmarks. Using the manually digitized cricothyroid membrane location as ground truth, we compare these four methods for accuracy and repeatability of identifying the landmark for cricothyrotomy, time of completion, and ease of use.  These preliminary results suggest that the accuracy of remote cricothyrotomy landmark identification is improved when the user is aided with visual and force cues. They also show that, with proper user initialization, landmark identification using remote palpation is feasible - therefore satisfying a key pre-requisite for future robotic solutions for remote cricothyrotomy.
\end{abstract}
\section{Introduction} \label{sec:intro}
\par Recent decades have seen rapid growth in robotic systems for a variety of surgical interventions. However, despite these advances, the use of robotic systems for trauma care remains relatively unexplored. There is a potential for robotic systems to perform life-saving procedures on wounded patients using a combination of telemanipulation and semi-automation. This work investigates one avenue to evaluate the feasibility of robotics in this space.
\par The use of robotics for trauma care can help in combat and disaster-related injuries. In Combat scenarios where it may be dangerous to send a human medic to stabilize and treat traumatic injuries, a robot can replace sending a human in harm's way. Similarly, in disaster scenarios where humans may have difficulty in accessing individuals in unsafe environments or in difficult-to-reach environments, robots may be able to reach areas not safely accessible to physicians within a short time. Also, developing the fundamental research for trauma-capable robots may allow a more ubiquitous presence of trauma-care robots in hospitals, ambulances, or in public places for rapid response to emergency situations.
\begin{figure}[htbp]
  \centering
  \includegraphics[width=0.65\columnwidth]{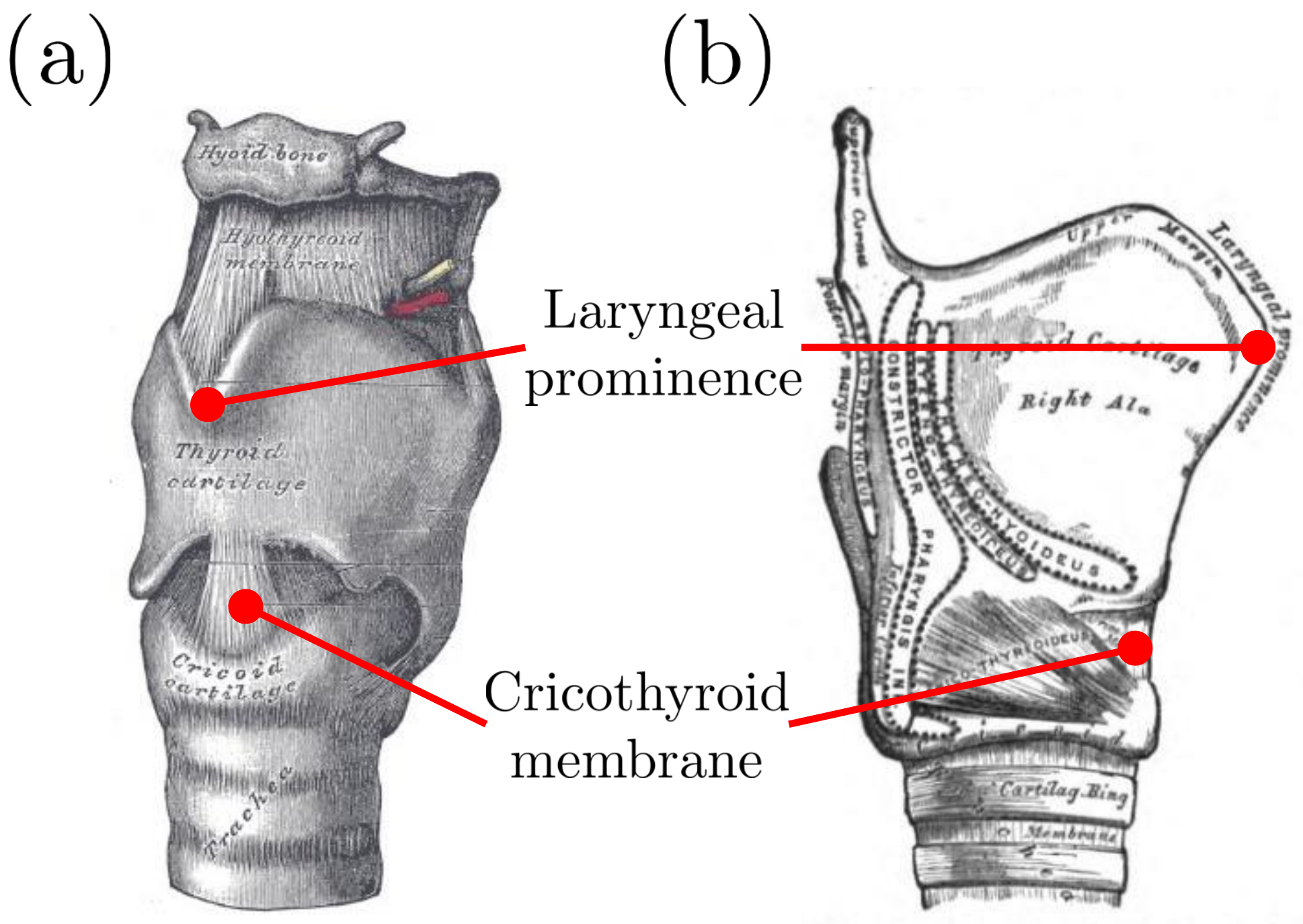}
  \caption{(a) anterior and lateral view of the larynx that shows the cricothyroid membrane. This membrane connects the contiguous front outer edges of the thyroid and cricoid cartilages. (b) side view of the human larynx. (\textit{Images adapted from Gray’s anatomy~\cite{gray1878}, plates 951 and 957})}\label{fig:cricothyroid_anatomy}
\end{figure}
\par One such key emergency procedure is Cricothyrotomy, which is a life-saving procedure to establish an airway in patients when standard intubation techniques are unsuccessful or unable to be attempted. To enable ventilation via cricothyrotomy, an incision in the cricothyroid membrane (Fig. (\ref{fig:cricothyroid_anatomy}a)) is used to provide access for an endotracheal tube. Identifying the correct incision landmark requires visual inspection of the laryngeal prominence (Adam's apple)~(Fig. (\ref{fig:cricothyroid_anatomy}a)) and concurrent manual palpation of the cricothyroid membrane. The procedure generally entails the stabilization of the trachea with one hand while palpating the laryngeal prominence down to the cricothyroid membrane with other. After this membrane is located, a scalpel is used to incise through the cricothyroid membrane to facilitate access to the airway. These are the key steps of the procedure prior to inserting the endotracheal tube. The procedure is concluded by securing the endotracheal tube using sutures or a separate fixation device. While the traditional approach to surgical cricothyrotomy uses a scalpel and the insertion of a finger into the airway, a robotic approach could leverage a cricothyrotomy kit (e.g. Cric-Key, Plumodyne$^{\circledR}$, Indianapolis, IN), which does not require excessive scalpel use and is made for simpler and controlled airway access in high-stress situations without compromising accuracy.
\par Identification of the cricothyrotomy landmark is critical for establishing an external airway path. Failure comes with a risk of severing the cricothyroid arteries, tearing the tracheal cartilage, damaging the trachea, or creating a false passage~\cite{erlandson1989cricothyrotomy, hart2010emergency}. The reported complication rates of the emergency cricothyrotomy range from 10\% to 40\%, primarily due to inaccurate landmark identification~\cite{hart2010emergency, D'Auria2013sim, Elliott2010accuracy, mace1988cricothyrotomy}. Cricothyrotomy is performed in life-or-death situations  and when other airway management options are contraindicated. Due to the rarity of the procedure, factors like procedural unfamiliarity and the limited knowledge of anatomy contribute to these above-mentioned complication rates, especially, in the tactical battlefield environments~\cite{Kotora2011}.
\par The goal of this investigation, at this stage, is to investigate the feasibility and repeatability of robot-assisted landmark identification in a best case scenario where the airway is assumed to be stabilized by a second hand or by a mechanical means. Once this feasibility is ascertained, investigations of possible use of robotic aids to reduce complications during Cricothyrotomy may be warranted. After reviewing the related works in section (\ref{sec:related_works}), section (\ref{sec:experiment_setup}) summarizes the experimental setup; section (\ref{sec:control_design}) discusses the control design used for the robot to accomplish the experiments; and  sections (\ref{sec:assistive_modes}) and (\ref{sec:telemanipulated_user_study}) discuss the different telemanipulation modes and the experimental evaluation, respectively.
\section{Related Works} \label{sec:related_works}
\par Related works to this investigation include works on mechanical imaging of tumors~\cite{Egorov2008,Althoefer2010,Patel2010}, mechanical probing~\cite{OkamuraGoldberg2015}, mechanical excitation for tissue impedance characterization~\cite{Goldman2013}, and robot-assisted palpation~\cite{Chalasani2016}. These works predominantly focused on the identification of tumors and have not been applied to palpation of the airways. Within the context of robot-assisted palpation for cricothyrotomy landmark identification, we believe this work to be probably the first to address feasibility and expected performance.
\par Specifically, there are research works that investigate the means of improving the performance for Cricothyrotomy that do not involve robot assistance. These attempts focus on improving cricothyrotomy training program either by introducing the simulators with some real-time feedback on performance~\cite{Bowyer2005hapticSim,D'Auria2013sim} or by filling up the gaps found in the existing procedural skill training programs~\cite{bennett2011crico}. These approaches demand the emergency physicians to be present on the site which may be prohibitively dangerous in certain scenarios of the battlefield and disaster rescue.
\par Within the broader scope of robotics for trauma care, works in orthopaedics~\cite{karuppiah2018robotics} for bone fixation and re-alignment have shown promise. The main exploration of robots for battlefield trauma has been the Trauma Pod~\cite{martinic2014glimpses}, which requires retrieval of the soldier prior to treatment whereas in our envisioned approach the robot would be used to offer a means for stabilizing injuries where conditions may not allow medics or robotic devices to achieve fast soldier recovery.  The key aspect of remote trauma care is telerobotics over long distances, which has been achieved through a variety of means (e.g.  ~\cite{lum2007telesurgery,harnett2008evaluation}). This area of research is otherwise unexplored.
\section{Experimental Setup}\label{sec:experiment_setup}
\begin{figure}
  \centering
  \includegraphics[width=\columnwidth]{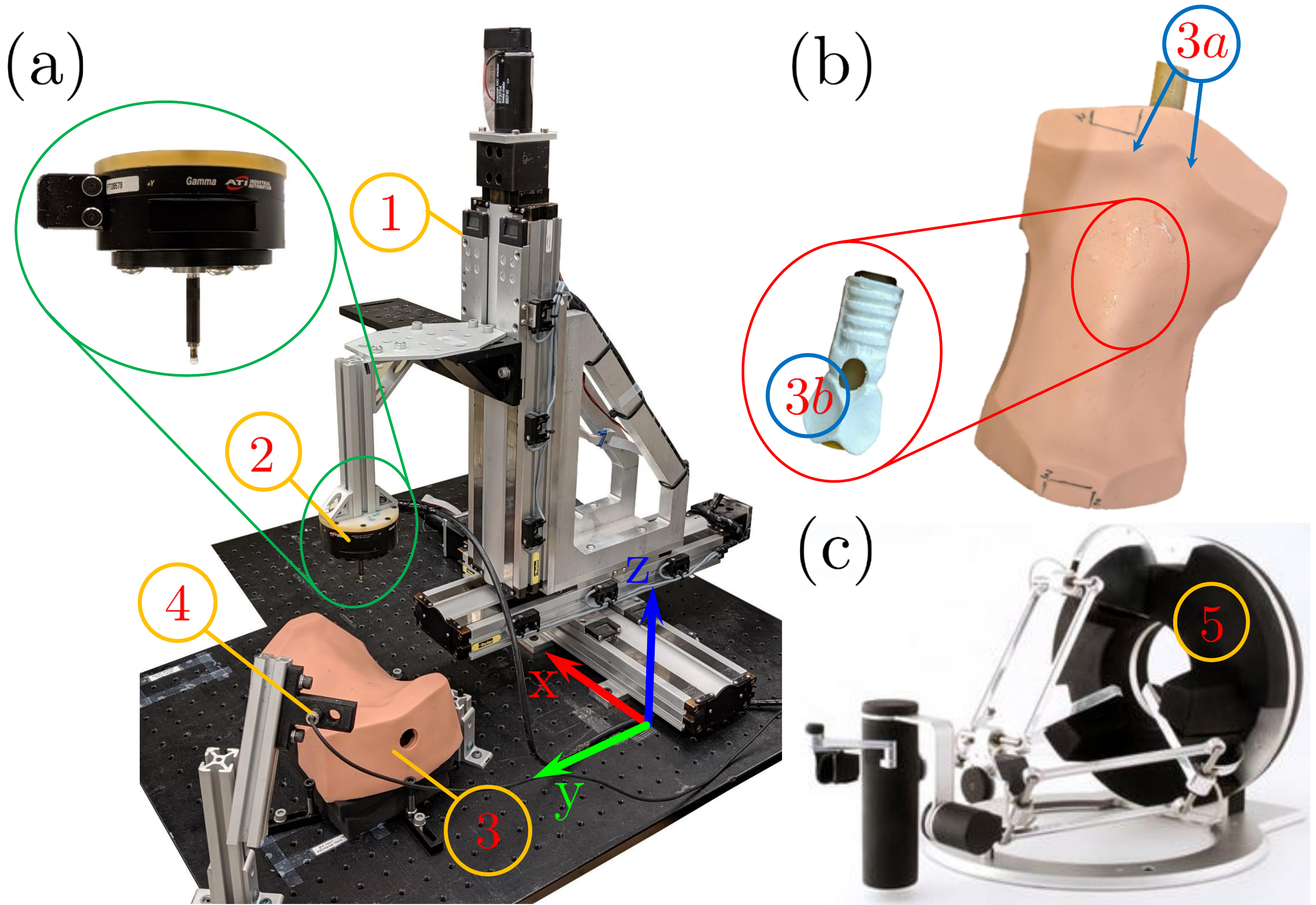}
  \caption{An experimental Setup (a) consisting of an XYZ cartesian robot {\large \textcircled{\small 1}} which manipulates the force sensor {\large \textcircled{\small 2}} to palpate the phantom throat anatomy {\large \textcircled{\small 3}}. A camera {\large \textcircled{\small 4}} captures the interaction between the probe and the phantom. (b) A trachea insert {\large \textcircled{\small 3b}} which has the opening at the location of the cricoid membrane. (c) A 7-DOF haptic device to telemanipulate the probe tip and to provide the force feedback to the user.}\label{fig:experimental_setup}
\end{figure}
\par Our feasibility experiments used a custom Cartesian robot with a distal force sensor and a Force Dimension Omega.7\textsuperscript\textregistered (Nyon, Switzerland) haptic interface as depicted in Fig.~(\ref{fig:experimental_setup}).  The robot is comprised of three orthogonally assembled linear stages (Parker\textsuperscript{TM} 404XR (Charlotte, NC)). At the distal end of the robot, an ATI (Apex, NC) Gamma\textsuperscript\textregistered~force/torque sensor was mounted and equipped with a plastic probe tip for palpation. The Omega.7 haptic interface is a high-fidelity device that can apply 3-axis force feedback on the user's hand while allowing 6-axis motion commands and a gripper control.
\par A Deluxe Cricothyrotomy Simulator from Simulaids (Saugerties, NY), shown in Fig. (\ref{fig:experimental_setup})(item {\large \textcircled{\small 3}}), was used as the throat phantom for the experiments. The phantom has a plastic base that mimics the throat anatomy from chin to collarbones. This cricothyrotomy training simulator used an adult-size trachea phantom (item {\large \textcircled{\small 3b}} in Fig. (\ref{fig:experimental_setup})), comprising the three major landmarks of the human larynx namely, thyroid cartilage, cricothyroid membrane, and cricoid cartilage. This trachea phantom was inserted into a recess in the plastic base in accordance with the instructions of the simulator. A soft rubber overlay skin was wrapped around the phantom to hold the trachea insert in place and cover the base. The recess for the trachea insert is large enough to mimic the movement of the trachea during the procedure. But for the experiments presented in this work, the trachea phantom was glued into the recess to avoid possible change in the landmark location while palpating. Neglecting the movement of the trachea enabled evaluating the accuracy and repeatability of the user-estimated locations of the cricothyroid membrane.
\par To provide the real-time visual feedback of the interaction between the user manipulated probe tip and the environment, a mono camera was mounted as shown in Fig. (\ref{fig:experimental_setup})(item {\large \textcircled{\small 4}}). The placement of the camera allowed an angled side view of the cricothyrotomy simulator in order to provide a reasonable sense of the depth of interaction between the probe tip and the palpated mock skin. The captured interaction was displayed on the 2D screen to the user. Also, a large opaque cloth was used as a blocking curtain between the robot side setup and the user side setup to block the direct visualization of the robot side setup.
\section{Control Design}\label{sec:control_design}
The Cartesian robot was controlled using a PD+Inverse Dynamics Low-Level Controller running at $1kHz$ on a real-time operating system. To control the forces in the direction of the surface normal while controlling tangential positioning, a hybrid force-motion controller was used. The torque, $\bs{\tau}$, applied to the joints of the robot is calculated as,
\begin{equation}\label{eq:pd_inv_dyn}
  \bs{\tau} = \bs{\tau}_m - \bs{\tau}_f + \mb{g} - \mb{B} \dot{\mb{p}}
\end{equation}
where, $\mb{B}$ is the joint-level estimated viscous friction, $\mb{g}$ is the gravitational force on the robot, torque $\bs{\tau}_m$ is the torque applied by the motion controller, torque $\bs{\tau}_f$ is the torque applied by the force controller, and $\dot{\mb{p}}$ is the linear velocity of the probe tip.
The motion controller torque is specified as,
\begin{equation}\label{eq:motion_control}
\bs{\tau}_{m}=\bs{\Omega}_m\mb{D}\ddot{\mb{p}}  \left[ \mb{K}_{p_m} (\mb{p}_{d}-\mb{p}) + \mb{K}_{d_m} (\dot{\mb{p}}_{d}-\dot{\mb{p}}) + \ddot{\mb{p}}_{d} \right]
\end{equation}
where, $\mb{D}$ is the mass matrix, $\mb{K}_{p_m}$ and $\mb{K}_{d_m}$ are diagonal positive-definite proportional and derivative feedback gain matrices, $\bs{\Omega}_m$ is a projection matrix that projects the output to the space corresponding with allowable motion directions, and $\ddot{\mb{p}}$ is the linear acceleration of the probe tip.
\par The force controller is specified as,
\begin{equation}\label{eq:force_control}
  \bs{\tau}_f =\bs{\Omega}_f \left(\mb{f}_{d} +\mb{K}_{p_f} \mb{f}_e +\mb{K}_{i_f} \int \mb{f}_e -  \mb{K}_v \dot{\mb{p}} \right)
\end{equation}
where, $\mb{K}_{p_f},~\mb{K}_{i_f}$ are the proportional and integral feedback gain matrices, $\mb{K}_v$ is an energy dissipation matrix for stabilizing the controller \cite{Khatib1987}  and $\bs{\Omega}_f$ is a projection matrix that projects forces into the space associated with constraint directions. The term $\mb{f}_d$ is a feedforward reference term for the desired applied force. The force error is defined as, $\mb{f}_e = \mb{f}_{d}-\mb{f}_{c}$. Also, the term $\bs{\tau}_f$ is negative in (\ref{eq:pd_inv_dyn}) because the desired force, $\mb{f}_d$, represents the force applied on the robot, not the force that the robot applies on the environment.
\par These force and motion projection matrices in (\ref{eq:force_control}) and (\ref{eq:motion_control}) are explicitly given by,
\begin{equation}\label{eq:force projection_matrix}
 \bs{\Omega}_f = \uvec{n}_e\T \uvec{n}_e
\end{equation}
\begin{equation}\label{eq:motion_projection_matrix}
 \bs{\Omega}_m = \mb{I}-\bs{\Omega}_f
\end{equation}
where, $\uvec{n}_e$ is the local environment surface normal and $\mb{I}\in\realfield{3\times3}$ is the identity matrix. The surface normal was estimated from contact force measurements using a moving average filter with 100 points. A generalization of how these matrices could be defined was presented in \cite{khatib1999general}.
\par For each of the telemanipulation modes, we used the motion controller until the probe tip made contact with the phantom skin. The force threshold for contact detection was $0.5N$. After detecting contact, the hybrid force controller was used to follow the user-induced motion commands while maintaining the constant force of $1.35N$ on the phantom.
\subsection{Bilateral Teleoperation for Force Feedback}\label{subsec:bilateral_teleop}
In this control mode, the Cartesian robot directly receives the reference position from the Omega.7. The reference command positions were scaled by $0.8$ telemanipulation gain and input into the motion controller in (\ref{eq:motion_control}). As the robot makes contact with the environment, interaction forces are sent back to the Omega.7 directly to relay those forces onto the user's hand. The gripper of the Omega.7 is used as a clutch; when the gripper is open, the user can move the Omega.7 freely without commanding the robot and without feeling interaction forces. When the gripper is closed, commands from the Omega.7 are sent to the robot and forces from the XYZ robot are relayed back onto the Omega.7 with an addition of the viscous damping terms.
\subsection{Stiffness Estimation}\label{subsec:stiffness_estimation}
For the modes that require motion excitation and automated palpation, the force controller in (\ref{eq:force_control}) was given a sinusoidal reference force along $\mb{z}$-axis (Fig. (\ref{fig:experimental_setup})). With this sinusoidal force input, it is possible to estimate the environment stiffness based on the most recent position and force data received from the robot.
\par
\begin{figure}
  \centering
  \includegraphics[width=\columnwidth]{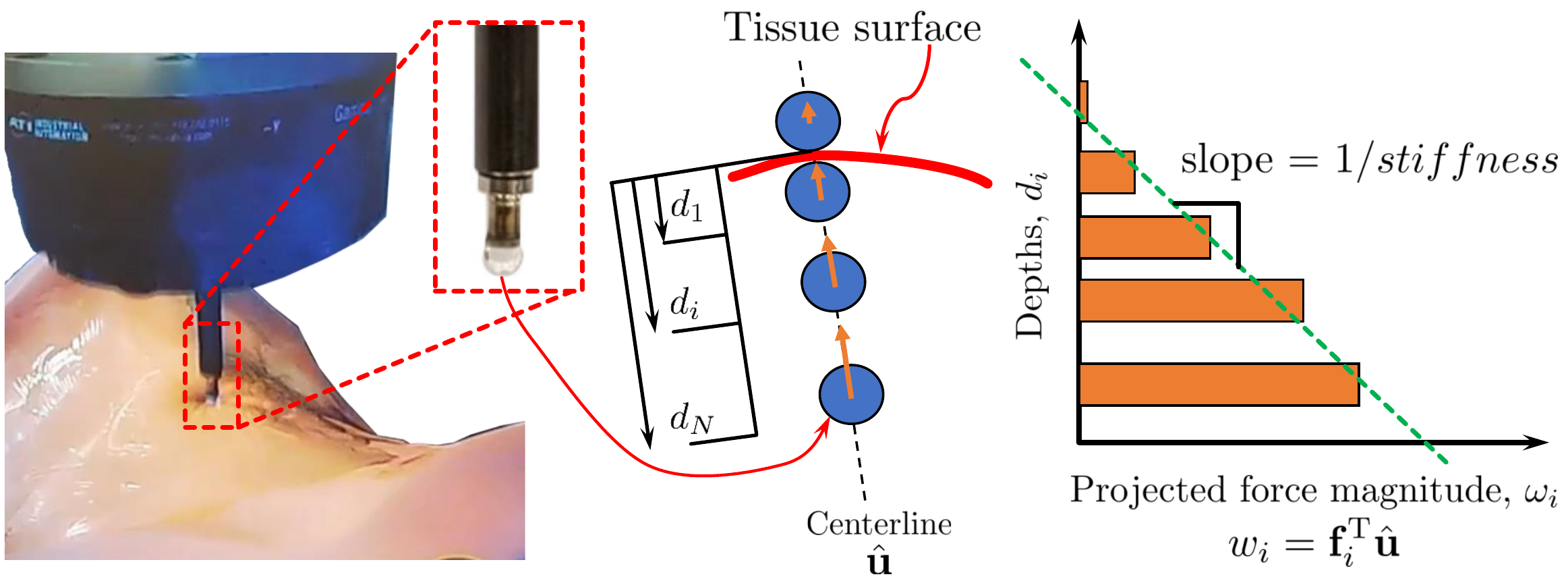}
  \caption{Stiffness estimation. The probe palpating the tissue (left), the calculated direction of motion and probing depth (middle), the stiffness calculation (right)}\label{fig:stiff_estimate}
\end{figure}

 For a given palpation cycle, a line is fit and regression is carried out on $N$ samples of both the forces and the depths of palpation. The line fit to the palpation motion is associated with a directional vector denoted by $\uvec{u}$ (Fig. (\ref{fig:stiff_estimate})). This vector designates the local tangent in the direction of end effector motion along the surface of the environment. The depths, $d_i$, are calculated for each probe tip position during the palpation cycle and is given by,
\begin{equation} \label{eq:depth_i}
  d_i = (\mb{p}_i-\bs{\mu}_{\mb{p}})\T \uvec{u}, \quad i = 1,2,...,N
\end{equation}
Here, $\bs{\mu}_{\mb{p}}$ is the mean of 250 most recent probe tip positions. The norm of the force, $\mb{f}_i$, projected along the directional vector $\uvec{u}$ is used to calculate the directional stiffness along the palpation direction. The projected norm at each position $\mb{p}_i$ is denoted as $w_i$ and is given by,
\begin{equation}
  w_i = \mb{f}_i\T \uvec{u}, \quad i = 1,2,...,N
\end{equation}
A line is fit through the pairs of data $d_i$ and $w_i$ where the slope of this line represents the directional stiffness of the environment for that palpation cycle at a particular location along the palpation path.
\par With the local directional stiffness of the environment being evaluated, a surface and the stiffness values over it were interpolated using the {\it{RegularizeData3D}} MATLAB function \cite{RegularizeData3D}. With the user commanded probe tip position, this surface was updated online and displayed on the screen as feedback for one of the telemanipulation modes described in section (\ref{sec:assistive_modes}). The repeatability of the stiffness estimation algorithm is not studied as the purpose here is to show the relative stiffness.
\subsection{Virtual Fixture}\label{subsec:virtual_fixture}
In this mode of operation, a virtual fixture was implemented on the Omega.7 side. To constrain the user's hand during telemanipulation, a forbidden region virtual fixture was created that disallows movement outside a certain box within the workspace.
\begin{align}\label{eq:vf}
f_i =
\begin{cases}
k(b_{i,max}-p_{i,d}) &,p_{i,d}>b_{i,max} \quad i = x,y \\
k(b_{i,min}-p_{i,d}) &,p_{i,d}<b_{i,min} \quad i = x,y
\end{cases}
\end{align}
where, $k=1 \frac{N}{mm}$ is the box wall stiffness, $p_{i,d}$ denotes the desired probe tip position coordinate, $b_{i,max}$ and $b_{i,min}$ defines the boundaries of the box based on the prior knowledge of the phantom location, and $f_i$ is the force transmitted to the user at the boundaries of the box. This mode of operation did not affect the force in the z-direction as the virtual fixture was only active after the probe tip had made contact with the mock skin. While exploring, when the user controls the probe tip to be inside the box, no force is applied, but as the user tries to exit the box, a linear constraint force pushes the user’s hand to keep the desired position within the confines of the virtual fixture.
\section{Control and Assistive Modes evaluated}\label{sec:assistive_modes}
\begin{figure}[htbp]
        \centering
        \includegraphics[width=0.75\columnwidth]{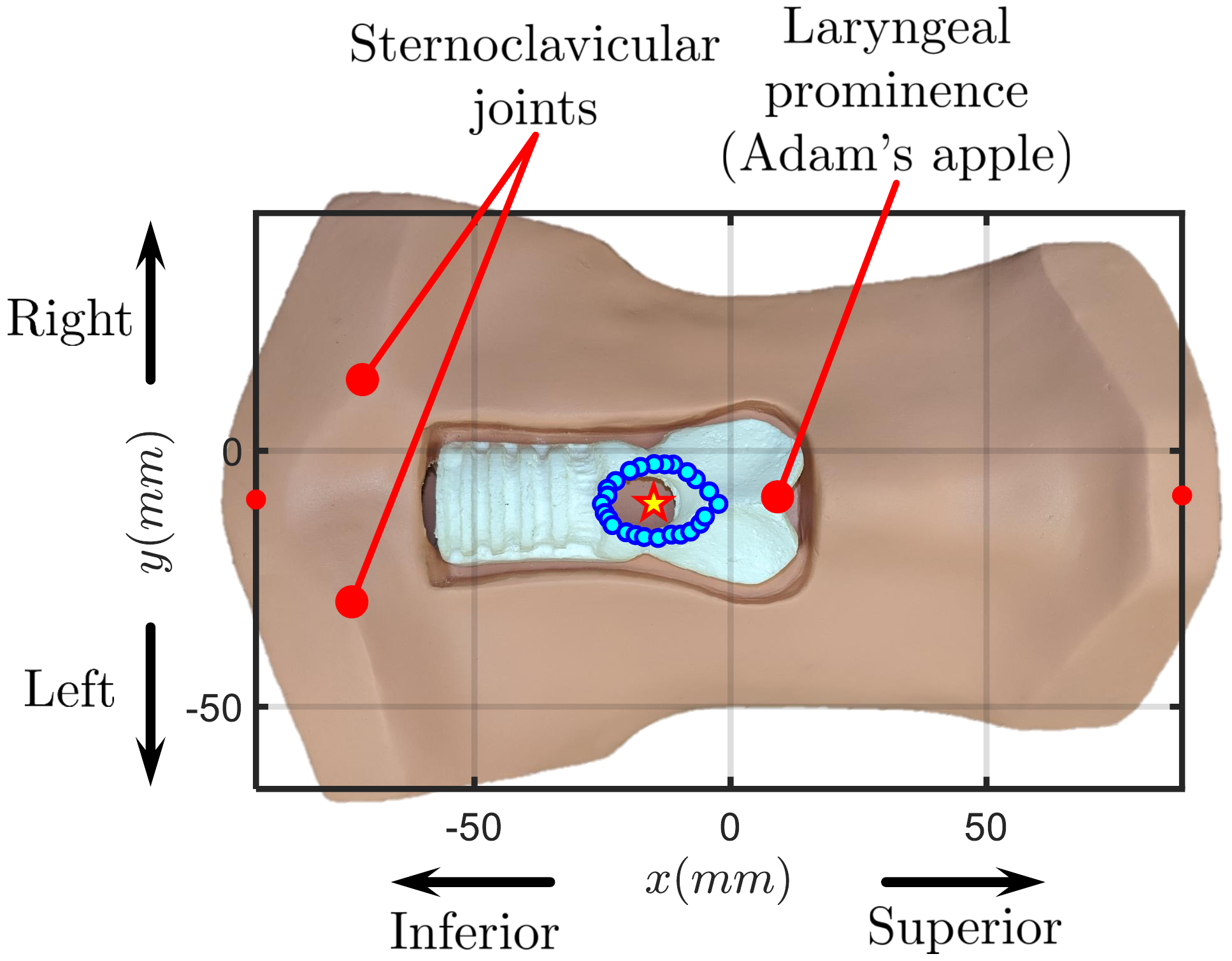}
        \caption{Blue circular markers depict the digitized points along the outer border of the cricothyroid membrane in trachea insert and the star marker represents the ground truth location of the center of the cricothyroid membrane}\label{fig:landmark_digitization}
\end{figure}

In this section, we discuss the telemanipulation modes that were evaluated for the feasibility study. These modes were evaluated for the accuracy and repeatability in the location of the landmark, the time of completion, and the ease of use. To evaluate the accuracy and repeatability, the location of the landmark was identified through manually digitizing the points along the circumference of the hole in the trachea depicting the cricothyroid membrane as shown in Fig. (\ref{fig:landmark_digitization}). The centroid of these points was used as a ground truth location of the center of the cricothyroid membrane.
\par
\begin{figure}[htbp]
    \centering
    \includegraphics[width=0.8\columnwidth]{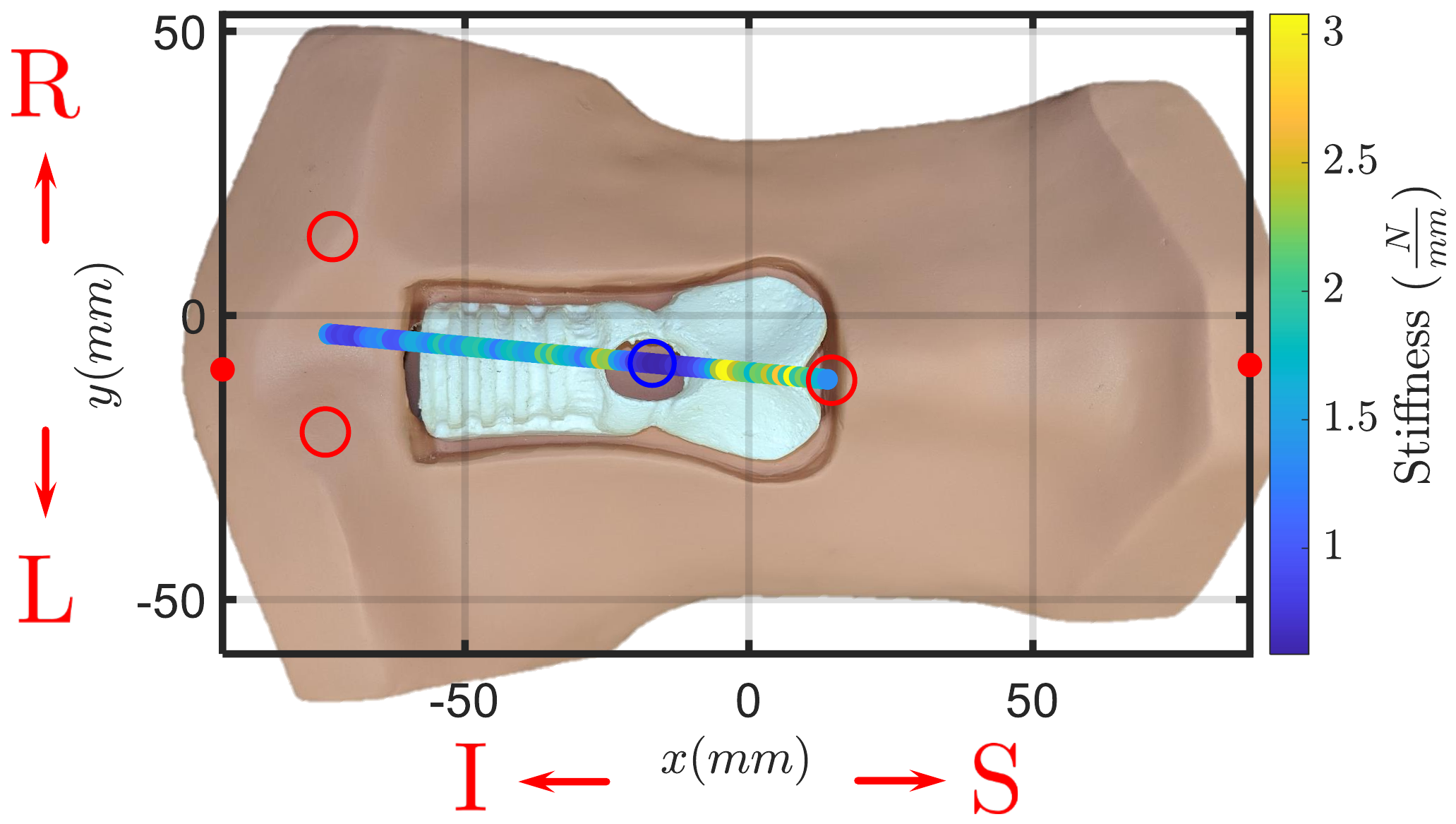}
    \caption{Representation of the procedure used to determine the location of the cricothyroid membrane in mode 4. The centers of the three red circles are the anatomical landmarks digitized by the user. The center of the blue circle is the location with the lowest stiffness along the line trajectory. The red letters on the axis labels are anatomical directions (S = Superior, I = Inferior, R = Right, L = Left)}\label{fig:sample_result_mode_4}
\end{figure}

\begin{figure}[htbp]
    \centering
    \includegraphics[width=0.92\columnwidth]{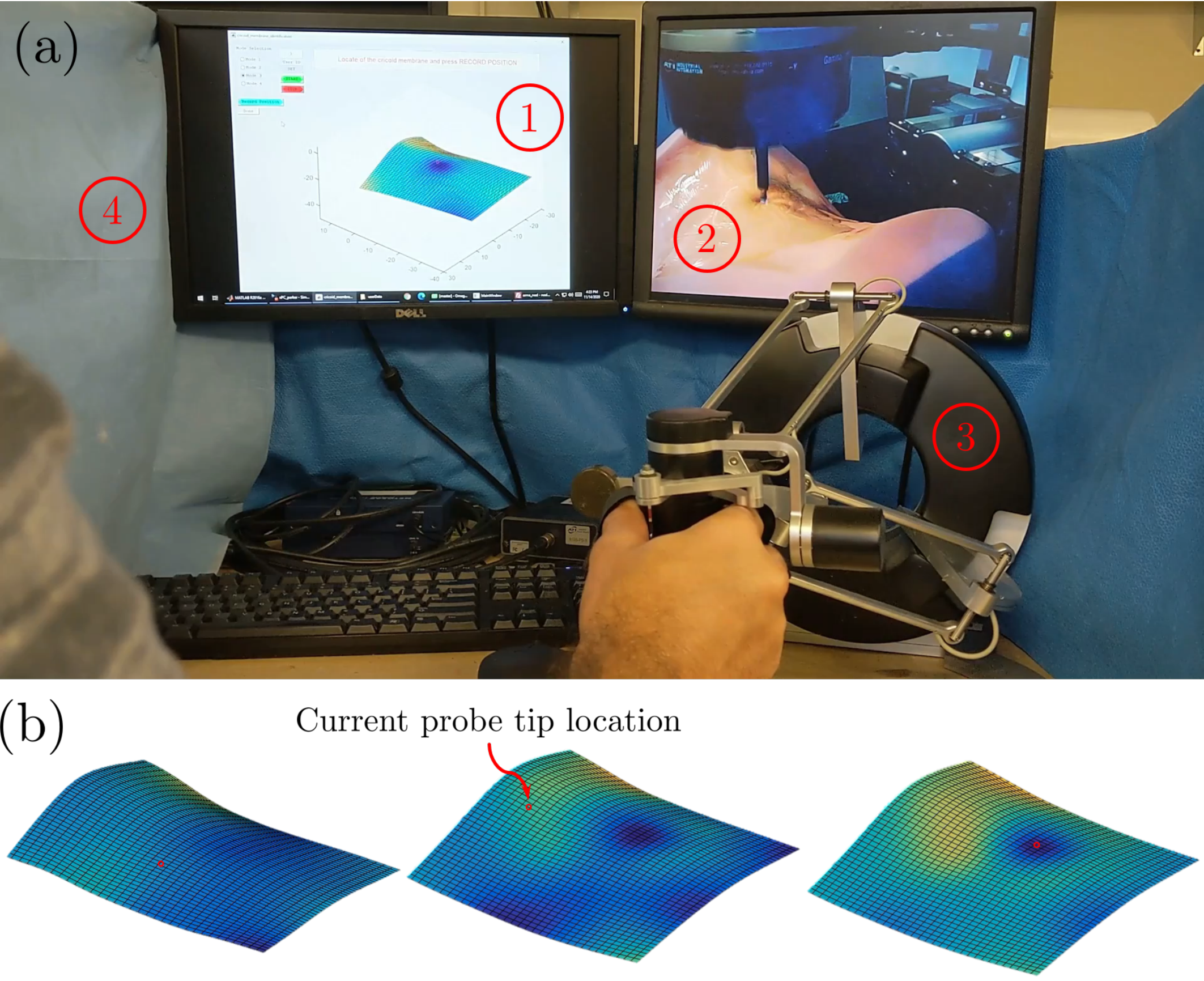}
    \caption{User side experimental setup (a) consisting of a 2D screen \textcircled{\small 1} to display the graphical user interface, another 2D \textcircled{\small 2} to display the live interaction between the probe tip and phantom, Omega.7 haptic device \textcircled{\small 3}, and the cloth \textcircled{\small 4} obstructing the direct visual access to the experimental setup. The time strip of the stiffness maps (b) at 3 different moments in mode 3. In these maps, regions with color closer to blue are softer than the rest}\label{fig:user_setup_mode_3}
\end{figure}

 Table (\ref{tbl:modes}) summarizes these telemanipulation assistive modes. The details of these modes are provided next.
\begin{table}[htbp]
\center
\setlength\tabcolsep{1.5pt} 
\renewcommand{\arraystretch}{1.4}
\begin{tabular}{ | c || p{6.2cm} | }
\hline
\textbf{Mode 1} & Visual display with no force feedback (FF) \\
\hline
\textbf{Mode 2} & Visual display with FF \\
\hline
\textbf{Mode 3} & Visual \& stiffness display with FF in lateral directions \& sinusoid excitation along z-direction \\
\hline
\textbf{Mode 4} & Semi automated with visual display \\
\hline
\end{tabular}
\caption{Assistive modes used in our evaluation study}\label{tbl:modes}
\end{table}
\begin{itemize}
\item \underline{Mode 1}: In this mode, the user telemanipulates the probe tip with the visual aid and without any force feedback. The visual aid was provided by streaming the real-time video of the site on a 2D screen. We used a virtual fixture as in \eqref{eq:vf} to bound the probe tip motion.
\item \underline{Mode 2}: In this mode, the user telemanipulates the probe tip with direct force feedback and visual aid. We used Bilateral teleoperation in  section~(\ref{subsec:bilateral_teleop}) to provide direct force feedback. We used a virtual fixture in section~(\ref{subsec:virtual_fixture}) to bound the probe tip motion. The visual aid was provided by streaming the real-time video of the site on a 2D screen.
\item \underline{Mode 3}: In this mode, the user telemanipulates the probe tip with visual aids and the direct lateral force feedback. We used superimposed sinusoid excitation in force which facilitated in computing local stiffness values online. We displayed an online updating stiffness map in addition to the real-time video of the site. The update rate for the stiffness map was $50Hz$. The amplitude, bias, and frequency of the sinusoid reference force, $(\mb{f}_d)_z$, are $1.8N$, $0.65N$, and $2Hz$ respectively. The maximum force during one time period of this sinusoid is comparable to the RMS force magnitude reported in \cite{Ryason2016} for cricothyrotomy on cadavers. Figure (\ref{fig:user_setup_mode_3}b) shows sample stiffness maps displayed to the user. The location of the probe tip is also shown on the stiffness map to help the user when digitizing the location of the \cric based on their judgment while looking at the stiffness map.
\item \underline{Mode 4}: Unlike the other three modes, this mode is automated and uses initial inputs from the user with the assumptions, (1) the user has clear enough visual feedback so that locating the visually explicit anatomical landmarks are trivial, and (2) the line joining the laryngeal prominence (Adam's Apple) and cricoid membrane (Fig. (\ref{fig:cricothyroid_anatomy}a)) lies in the Sagittal plane. Considering these assumptions, users were instructed to telemanipulate the probe tip to locate the sternoclavicular joints (Fig. (\ref{fig:experimental_setup})(item {\large \textcircled{\small 3a}})) and the laryngeal prominence (Fig. (\ref{fig:cricothyroid_anatomy}a)). Using these user estimated locations, a line trajectory was generated which passes through the mid-point of the line joining sternoclavicular joints and the laryngeal prominence. Using the superimposed sinusoid force excitation as mode 3 in z-axis, the robot followed the line trajectory where motion along the z-axis was controlled using the hybrid force-motion controller. The point with the lowest stiffness along the line (as shown in Fig. (\ref{fig:sample_result_mode_4})) was designated as the location of cricothyroid membrane.
\end{itemize}
\par The preliminary experimental evaluation included these modes of operation where each user repeated the experiment 10 times for performance evaluation as outlined in the following sections.
\section{Preliminary Experimental Evaluation and Results}\label{sec:telemanipulated_user_study}
Prior to conducting a user-study, three of the four co-authors used the system to evaluate the performance and feasibility of the  remote landmark identification for cricothyrotomy. Only one of the three co-authors had significant experience using the system, but has not gone through the entire experimental protocol prior to collecting data. All the users had a period of familiarization with using the system (approximately 10 minutes) and they carried the experiments with all the four modes of robot assistance in a randomized order among each user.
\par At the end of each mode except mode 4, users were instructed to bring the probe tip over to the location of the \cric landmark that they identified. Also, the total time to completion, from the point when robot control was given to users to the point they record their identified landmark location. Those estimated locations then are compared with the digitized ground truth location of the landmark as in Fig. (\ref{fig:landmark_digitization}).
\par
\begin{figure}[htbp]
  \centering
  \includegraphics[width=\columnwidth]{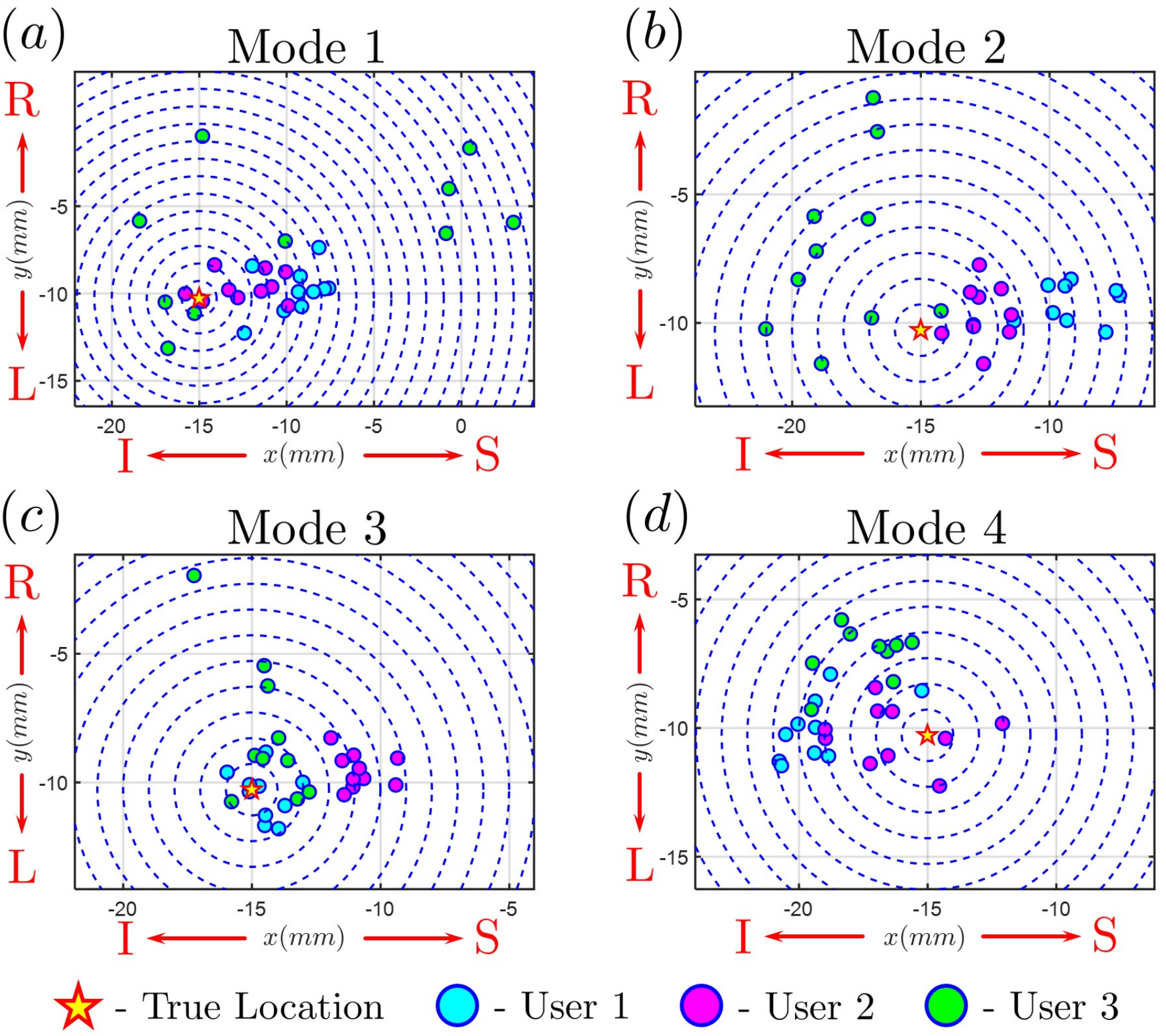}
  \caption{Polar plots of the user-estimated landmark locations depicted by the blue markers. Radial grid lines are drawn in 1mm steps. The red letters on the axis labels are anatomical directions (S = Superior, I = Inferior, R = Right, L = Left)}\label{fig:user_estimated_locations}
\end{figure}

 Figure (\ref{fig:user_estimated_locations}) shows the estimated locations of the cricothyroid membrane by each user in each mode. From the figure, it can be seen that modes 1 and 2 had the widest scatter in the data while mode 3 seemed to produce the closest clusters of data to the ground truth. Mode 4 was biased to the inferior direction due to the user's difficulty in discerning the true location of the laryngeal prominence when using only a single camera view. Table~(\ref{tbl:results}) summarizes the average, $\overline{(\cdot)}$, and standard deviation, $\sigma(\cdot)$, for the norm of location error and the completion time.
\begin{figure}[htbp]
  \centering
  \includegraphics[width=\columnwidth]{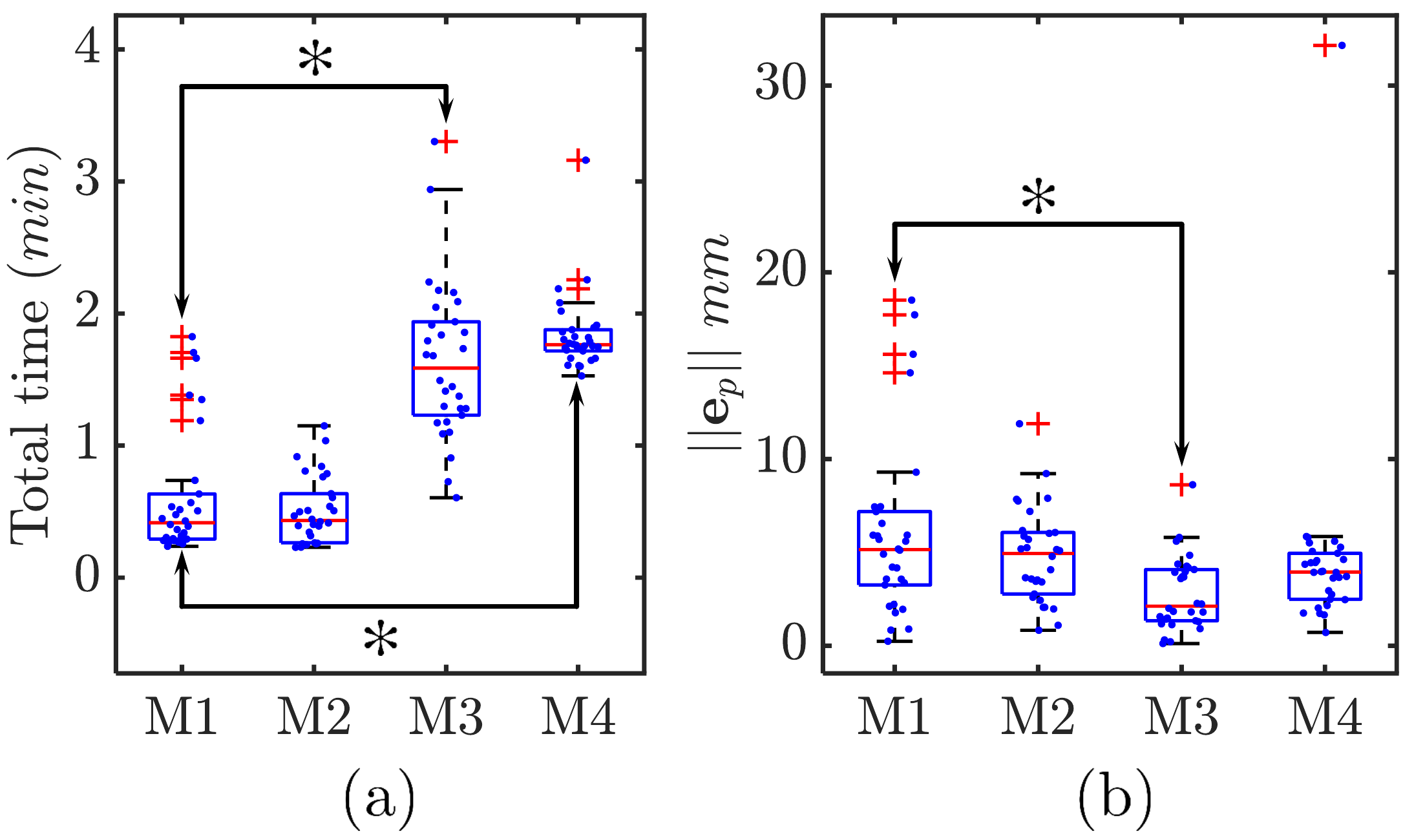}
  \caption{Box plots (a) total time to completion (b) norm of position error. The central red line is the median, with the box covering the 25th to 75th percentiles of the data. Outliers, plotted with a cross, are outside 2.7 standard deviations from the mean, assuming normality. Small dots represent all the data. Asterisk symbols designate a statistically significant difference in means.}\label{fig:box_plots}
\end{figure}
\begin{table}[htbp]
\center
\setlength\tabcolsep{1.5pt} 
\begin{tabular}{c||c|c|c|c}
& \textbf{Mode 1} & \textbf{Mode 2}  & \textbf{Mode 3} &  \textbf{Mode 4}\\ \hline \thickhline
$\overline{\lVert \mb{e}_p\rVert}, (\sigma(\mb{e}_p)) mm$ & 6.0, (4.8) &  4.8, (2.5)  &  2.8, (1.9)  &  4.7, (5.4)\\ \hline
$\overline{t}, (\sigma(t))$ min & 0.62, (0.48) & 0.51, (0.26) & 1.63, (0.59) & 1.84, (0.30)  \\ \hline
\end{tabular}
\caption{Mean and standard deviation of $\lVert \mb{e}_p\rVert$ and total time ($t$) for the assistive modes evaluated}\label{tbl:results}
\end{table}
\par Figure~(\ref{fig:box_plots}) shows box plots for the completion time and the norm of the position error $\lVert \mb{e}_p\rVert$ between the user-digitized location in each mode and the ground truth location. The results suggest that mode 3 required more time to identify the landmark location, but resulted in the least amount of error. The time to completion for mode 4 includes both time for identifying and recording the initial user inputs and completing the autonomous line scan. The time reported in this mode is not directly comparable to the results from the other modes as the time required to complete the autonomous scan depends on the length of the computed line path which in turn depends on the participant's anatomical knowledge and also on the maximum speed allowed when moving in the air (which can be significantly improved).
\par A comparison of the position error means relative to Mode 1 reveals that only mode 3 has a statistically significant difference of means with a $p$-value of 0.001 (two-sided t-test with $\alpha=0.05$). A comparison of time of completion means relative to Mode 1 reveals that only modes 3 and 4 have a statistically significant difference in means relative to Mode 1 (two-sided t-test with $\alpha=0.05$) and both had $p$-value of less than 0.001.
\section{Discussion}\label{sec:discussion}
\subsection{Observations about the results}
The incision for the \cric is considered acceptable if it does not injure the anatomy around the cricothyroid membrane. The accuracy and repeatability results shown here consider the center of the cricothyroid membrane for evaluation. This indicates that the acceptable error in identifying the location depends on the average size of the cricothyroid membrane. As in \cite{Nutbeam2017}, for the western population, the size (Width-Height) of cricothyroid membrane reported in males is ($13mm-10mm$) and in female, it is ($10.5mm-7.5mm$). Considering this information, the errors reported in mode 1 indicate the possible danger of relying on the visual information only. Table~\ref{tbl:percentages} shows the percentages of estimated locations that fall within the average size of a male/female cricothyroid membrane. The size of our training phantom model was closer to the male anatomy in size and these percentages should be only used for observing trends since our model was not exactly equal in size to the average male anatomy. Nevertheless, the reduced trends in \cric location estimation error as observed in Table~\ref{tbl:results} and Table~\ref{tbl:percentages} support the conclusion that modes 3 and 4 can provide higher accuracy and repeatability compared to an unassisted telemanipulation mode or a bilateral telemanipulation mode with force feedback.
\begin{table}[htbp]
\center
\setlength\tabcolsep{6pt} 
\begin{tabular}{c||c|c|c|c}
  & \textbf{Mode 1} & \textbf{Mode 2}  & \textbf{Mode 3} &  \textbf{Mode 4}\\ \hline \thickhline
Male &~$56.67\%~$ &~$66.67\%~$ &~$90\%~$ &~$90\%~$ \\ \hline
Female &~$43.33\%~$ &~$46.67\%~$ &~$73.33\%~$ &~$60\%~$ \\ \hline
\end{tabular}
\caption{Percentages of estimated locations that fall within an average male/female cricothyroid membrane for each mode}\label{tbl:percentages}
\end{table}
\subsection{Study limitations and improvements }\label{sec:study_assumptions_improvements}
\par This work presents a preliminary investigation of the feasibility of remote landmark identification for cricothyrotomy. It is limited in scope to answering feasibility questions about remote identification of the cricothyrotomy site when using a telemanipulation interface with and without assistive behaviors. Successful future deployment of robot-assisted cricothyrotomy would require much more than this step of palpation and landmark identification. For example, there is a need to stabilize the trachea when carrying out the access incision. There is a need for robotic embodiments that would allow dual-arm operation with tool exchange and with integrated sensing. All of these are key steps to achieving on-site battlefield trauma assessment and airway stabilization, but the successful identification of the landmark for cricothyrotomy is the prerequisite critical step for any successful future deployment of robotic solutions to this problem and that is why this work is focused on exploring landmark identification.
\par The semi-automated option (Mode 4) assumes that the patient's neck is set straight and untwisted. This is a rather limiting assumption that would make the use of this mode dangerous without intelligent pose estimation and visual detection of anatomical landmarks.
\subsection{Observational comments }\label{sec:observational_comments}
\par Even for landmark identification, there are some uncertainties that have to be addressed so that the robot-assisted approach can be adaptable to any possible scenarios for this procedure. The anatomical uncertainties depend on the age of the victim, palpability of the neck, and the type of injury. The age of the victim plays an important role as the anatomical landmarks which helps to locate the cricoid membrane are not fully developed in the victims of young age. Also, the cricoid cartilage is the narrowest portion of the airway for young victims which makes the task of landmark identification more challenging for manual palpation. In such scenarios, the robot-assistance comes handy as different probes can be used to conduct the robotic palpation. The palpability becomes a criterion considering the patients who have excessive neck swelling due to injury or excessive fat accumulation at the neck. For these patients, the emergency cricothyrotomy procedure involves other tools like spreading clamps, identifying other key anatomical landmarks like the hyoid bone \cite{Robert1983}, or cutting through the fat to access the underlying anatomy for palpation. The injury to the parts of the body other than the airway might not allow stabilizing the victim in the right posture required for the \cric procedure. As in this study, injury to the anatomy used as an aid in determining the landmark can contraindicate the use of such a method even though its better performance otherwise.
\section{Conclusion}\label{sec:conclusion}
In this paper, we showed the feasibility of robotic palpation methods for the detection of the cricothyroid membrane as the first step in robot-assisted cricothyrotomy. Several modes of telemanipulation were evaluated and the results suggest that relying on visual feedback alone (Mode 1) can be associated with unacceptably large errors. Introducing the force feedback (Mode 2) showed improvement but the accuracy and repeatability are more promising when the online updating stiffness map is displayed (Mode 3) in addition to visual and force feedback. Automating the scanning task (Mode 4) yields comparable accuracy and repeatability as mode 3. The total time to completion can be improved substantially using the visual feedback that provides a better perception of depth and increasing the maximum speed allowed while moving in the air and scanning the anatomy.
\bibliographystyle{IEEEtran}
\bibliography{main}
\end{document}